\documentclass[11pt,a4paper]{article}
\usepackage[hyperref]{acl2018}
\usepackage{latexsym}
\usepackage{times}
\usepackage{latexsym}
\usepackage{todonotes}

\usepackage{url}

\aclfinalcopy 

\title{Discourse Embellishment Using a Deep Encoder-Decoder Network}

\author{Leonid Berov\\
  Institute of Cognitive Science \\
  University of Osnabr\"uck, Germany \\
  {\tt \url{lberov@uos.de}} \\\And
  Kai Standvoss \\
  Donders Institute for Brain, Cognition \\and Behaviour\\
  Radboud University, The Netherlands\\
  {\tt \url{kstandvoss@uos.de}} \\}

\date{}

\begin{document}
\maketitle
\begin{abstract}
We suggest a new NLG task in the context of the discourse generation pipeline of computational storytelling systems.
This task, textual embellishment, is defined by taking a text as input and generating a semantically equivalent output with increased lexical and syntactic complexity.
Ideally, this would allow the authors of computational storytellers to implement just lightweight NLG systems and use a domain-independent embellishment module to translate its output into more literary text.
We present promising first results on this task using LSTM Encoder-Decoder networks trained on the WikiLarge dataset.
Furthermore, we introduce ``Compiled Computer Tales'', a corpus of computationally generated stories, that can be used to test the capabilities of embellishment algorithms.
\end{abstract}

\section{Introduction}
Narratives can be analyzed to consist of at least two layers: plot---what is told---and discourse---how it is told.
Usually, computational systems first generate the events of the plot, and then decide how to render these in text~\cite{gervas2009}.
This means that a tight coupling exists between the knowledge bases that are used for plot generation, and the NLG modules for discourse rendering.
The result is that it is hard to share these modules between systems, while the implementation of a custom module to generate literary discourse requires significant effort and linguistic expertise.
Indeed, our subjective feeling is that often it is possible to recognize repeating textual patterns when reading several stories created by the same system.

To alleviate this problems we suggest a new NLG task, \textit{textual embellishment} (TE), with the goal to automatically increase the lexical and syntactic complexity of a text.
Textual embellishment can be understood as the inverse task to textual simplification (TS), which has been researched for at least two decades~\cite{shardlow2014}.
Recent TS systems are trained on vast corpora using machine learning techniques and work independent of domains or handcrafted rules.
These results open the intriguing possibility for an equally general resource for TE.\ 
This could allow the authors of storytelling systems to safe time by implementing only thin NLG modules and perform subsequent embellishment.
We would like to strongly caution that TE most likely will prove to be a more difficult task then TS.
While the latter typically results in removing information in a text, the former might lead to an automated adding of information which can introduce semantic contradictions.
However, on a syntactic level TE ideally just results in periphrastic constructions with no added information.

\section{Related Work}
Traditionally, narrative generation systems like \citet{meehan1977tale,lebowitz1985story} were devised to generate macro-level plot and character elements without connecting them into a comprehensive story in natural language. 
Later work has sought to combine plot and NLG into a unified system~\cite{callaway2002}. These full pipelines generate story elements and employ rule-based language models that produce naturally sounding narratives. While these approaches make use of text embellishment, the disadvantage of these rule-based approaches is that rules for text enrichment have to be conceived of in advance by the system's architect. The work presented here, in contrast, seeks to learn a natural language representation from human data to automatically embellish priorly generated narratives. 

This approach can be located in the field of natural language processing. Within this area of research several related subfields can be identified, including text summarisation, statistical machine translation, and most notably, TS. Both, text summarisation and simplification, seek to extract the relevant information of a phrase and lose all linguistic embellishment that is deemed unnecessary for its understanding. Our work has a complementary objective, which is enriching text while maintaining its original meaning. This generative task is related to recent developments in computer vision using Generative Adversarial Networks for problems like image super resolution \cite{ledig2017photo}. Therein similar challenges are faced as new information has to be ``hallucinated'', requiring particular caution in content sensitive domains. 

Within TS, \citet{shardlow2014} distinguishes between syntactic and lexical simplification. While the former approach seeks to facilitate readability by reducing the complexity of sentence structure, the latter seeks to replace words deemed difficult for a target audience by words that are easier understandable. Previous work has focused on both tasks individually~\cite{siddharthan2006syntactic,paetzold2017}, while recent approaches addressed syntactic and lexical simplification simultaneously \cite{wang2016,zhang2017}. These latter systems cast text simplification as a monolingual machine translation problem and borrow insights from automatic natural language generation~\cite{wen2015semantically}. Therein, the task of simplifying text is framed as a translation problem between ``complex English'' and ``simple English''. The recent success of Deep Learning systems in ``neural machine translation''~\cite{bahdanau2014neural,cho2014learning} has increased the employment, specifically of the LSTM encoder-decoder architecture~\cite{sutskever2014}. These models make use of the Long Short-Term Memory neural network architecture \cite{hochreiter1997long} to solve a sequence to sequence task. Therein, a mapping from a \emph{source} sequence to a \emph{target} sequence is learned both of which may be of variable length
. To that goal, an \emph{encoder} LSTM generates a sequence of hidden representations using the source, which is then used by a second \emph{decoder} LSTM to generate the target sequence word by word, whereof each is conditioned on all previous outputs.  Here, we follow the same approach but address the opposite problem to TS, translating from ``simple'' to ``complex'' English, building on the success of previous neural network approaches. 

\section{Discourse Embellishment}
For the trained network, two LSTM layers per encoder and decoder were chosen, since \citet{wang2016} demonstrated that this is sufficient for sorting, reversing and reordering operations in a textual simplification task.
The network was set up and trained using Harvard's OpenNMT-tf framework~\cite{klein2017}.
Parameters were selected as suggested by~\citet{zhang2017}: Each layer contained 512 hidden units with weights uniformly initialized to [-0.1, 0.1], learning was performed with a 0.2 dropout rate~\cite{zaremba2014} and Luong attention was used by the decoder~\cite{luong2015}.
If not stated otherwise, training was always performed for 24 epochs, with a learning rate of 1 and a simple gradient descent optimization, where gradients where clipped when their norm exceeded 5~\cite{pascanu2013}.
After 18 epochs a decay rate of 0.25 was applied each epoch.
Furthermore, pre-trained 300 dimensional GloVe vectors were used to initialize word embeddings~\cite{pennington2014}, and the vocabulary size was 50,000.

\subsection{Story Corpus}
The performance of a discourse embellishment system should ideally be judged on domain-specific texts.
To the best of our knowledge no corpus of computationally generated stories has been compiled so far.
Since no common resources for NLG generation have been deployed, the stories that have been generated by different systems over the decades differ markedly in their language.

To reflect this diversity we decided for a breadth-focused approach while setting up \textit{Compiled Computer Tales} (CCT), a corpus of computationally generated stories.
Since the aim of this corpus is to be used as a qualitative test set, and not as training data for machine learning algorithms, we collected at most three stories from as many systems as available to us, instead of compiling many stories from few systems.
Because it is unfeasible to deploy all individual systems, we instead opted for using stories that have been reported in scientific publications, although this runs the danger of biasing the corpus towards high-quality exemplars.\interfootnotelinepenalty=10000\footnote{
A notable exception is ``Simlextric''~\cite{riegl2018}, which is accessible online and was used to generate novel stories for the corpus.}
The resulting corpus can be found online\footnote{\url{https://github.com/cartisan/CompiledComputerTales}} and we cordially invite researchers to expand this necessarily incomplete collection.

At the moment of writing, it contains 14 stories from 8 storytellers, ranging from early systems like ``Novel Writer''~\cite{klein1973} to recent ones like ``Simlextric''~\cite{riegl2018}.
It is formatted in a way that allows separation based on system, story, paragraph (45 in total) or sentence (290 in total); depending on the task at hand.
We also provide a python script that can be used to perform such splits, as well as common pre-processing steps like tokenization or named entity anonymization.

\subsection{Sentence Based: Lexical Embellishment}\label{sec:sen-bas}
As with the more common TS task, we trained our network on a sentence-aligned dataset.
The largest simplification dataset is, to the best of our knowledge, WikiLarge, which contains 296,402 automatically aligned sentences from the ordinary and simple English Wikipedias in its training set.
To reduce vocabulary, named entities in the dataset were anonymized~\cite{zhang2017}.

Unlike for simplification, for the present task the network was trained to generate original ('complex') English based on simple English input.
The performance of the system is evaluated using the common BLEU metric~\cite{papineni2002}, which measures the degree to which generated embellished text differs from the language employed in the original Wikipedia.
Training is continuously evaluated on a held-out evaluation set using the Moses toolkit~\cite{koehn2007}, and BLEU stagnates after around 18 epochs.
The final performance on the held-out test set is a BLEU of 56.02 on the tokenized data, which is a reasonable performance\footnote{
The state of the art system from 2014 achieves a worse BLEU of 48.97 on the considerably easier textual simplification task using this dataset.
However, current state of the art strongly outperforms this with a BLEU of 88.85, again, on the simpler text simplification task~\cite{zhang2017}.
}.

To judge the performance of this model on CCT, the corpus had to be split in sentences, tokenized, and named entities had to be anonymized in a manner comparable to WikiLarge.
Since no gold-standard of embellished tales exists, quantitative evaluation of embellishment quality is not possible.
Comparing the embellished version with the original ones shows that the network mostly learned to correctly reproduce its input (BLEU 92.13).
If differences that are due only to out-of-vocabulary words and minor formatting differences are ignored, then roughly 83\% of the sentences are simply reproductions of the input.
Around 5\% of the generated sentences do not bear any resemblance to the corresponding input, which we consider to happen when the network overfitted a certain type of input.
2\% of the generated sentences miss individual words.
The remaining 10\% contain syntactically correct lexical substitutions, which can be considered cases of lexical embellishment.
Some examples of felicitous embellishment are the mappings: \textit{
    Because of this $\rightarrow$ consequently $\|$ 
    tried $\rightarrow$ attempted $\|$
    wanted to $\rightarrow$ sought to $\|$
    bossy $\rightarrow$ overbearing};
while examples of failed embellishment are: ~\textit{
     sinful $\rightarrow$ drinking $\|$
     lioness $\rightarrow$ cat $\|$
     ubiquitous $\rightarrow$ familiar $\|$
     towards princess $\rightarrow$ towards LOCATION@1}.

\subsection{Paragraph Based: Degradation}
Sentence aligned training enabled the network to perform lexical embellishment, however, no syntactical embellishment like concatenation or parenthetical insertion was observed.
It stands to reason that such sentence-level reordering can be better picked up by learning from a corpus that aligns whole paragraphs, where several sentence are simplified/embellished at a time.
A promising candidate is the document aligned Wikipedia corpus compiled by~\citet{kauchak2013}.
After the necessary preprocessing and splitting steps the resulting dataset contained 134,233 paragraphs in the training set.

Training a network on this dataset using the same setup was not successful.
The model achieved a BLEU score of 3.55 on the held-out test set, which is a clear sign of its failure.

A manual inspection of the dataset revealed, that the employed alignment of paragraphs is rarely meaningful, since paragraphs of the same ordinal number often fail to contain semantically related information, let alone simplified/embellished versions of each other.
In fact, further literature review did not reveal work on either paragraph or document aligned simplification, as well as no published results using the document aligned dataset. 
This lead us to suspect that the encountered problems are not of trivial nature, and abandon this direction of inquiry.

\subsection{Pair Based: Syntactical Embellishment}\label{sec:sen-pair}
An inspection of the WikiLarge dataset revealed that the sought-for syntactical reorderings can also be found there.
In particular, the sentence-alignment does not necessarily result in a bijective mapping from complex to simple sentences, but sometimes in a 1:2 mapping, when complex sentences are split in two.
This lead us to the hypothesis that the network described in section~\ref{sec:sen-bas} should already be able to perform syntactical embellishment, when presented with two input sentences.

To test this we pre-processed the CCT corpus by concatenating, with a space, two consecutive sentences.
This resulted in sentence-pairs that in most cases had the potential to be combined, because they shared subjects or were causally related.
In some cases, syntactical combinations were not meaningful, especially when sentence pairs crossed paragraph or story boundaries as a result of the unsophisticated concatenation operation.

Comparing the embellished version with the original ones shows that the network is doing an acceptable job in keeping original and embellishment related (BLEU 68.20).
Approximately 56\% of the generated pairs were combined by the network into one sentence.
In 42\% of the cases the network did not combine the pairs, in which case it usually kept the first sentence and removed the second.
In the remaining 2\% the networks output was not related to the input.

The combination of sentences was performed by the network in different ways.
Most commonly, it conjoined them using a comma and the conjunction \textit{and}, or replaced the period sign with a comma, while changing the following letter to lower case.
We also observed cases where the pronouns \textit{who, which} or \textit{where} were employed.
Another mode of combining sentences employed by the network was by using participles:
\textit{\ldots she robbed PERSON@2 of her illusions . She said : '' PERSON@3 took \ldots ''} $\rightarrow$ \textit{\ldots she robbed PERSON@2 of her illusions \textbf{, saying} : '' PERSON@3 took \ldots ''}.
The most interesting case appeared to involve an inclusion: \textit{PERSON@1 saw the affair . PERSON@1 was jealous} $\rightarrow$ \textit{PERSON@1 \textbf{was jealous of} the affair}.
In a previous run the same sentence was embellished using nominalisation: \textit{PERSON@1 saw the affair \textbf{with jealousy}}.
For more context, a comparison between an original and a pair-embellished story can be found in the appendix.

\section{Conclusion}
NLG systems in computational storytellers are commonly dependent on the systems' knowledge base and require idiosyncratic, hand-coded rules if complex language output is desired.
Based on this observation we propose to extend the discourse generation pipeline with a TE step that takes as input simple English sentences and performs a monolingual translation into lexically and syntactically more complex English. 
This approach has the benefit of allowing the authors of storytellers to save time by implementing only simple custom text generation modules, and employ a domain-independent embellishment module to translate into more sophisticated text.

The LSTM Encoder-Decoder approach explored in this paper is not ready for productive use as is, mainly because it does not always fail gracefully, i.e.\ produces well-formed and semantically appropriate language when embellishment fails.
However, it shows promising results by demonstrating that the same trained network is capable of producing interesting lexical as well as syntactical embellishment of varying kind.

We are confident that the presented results can be improved by employing techniques that have proven beneficial in a simplification setting, like e.g. the reinforcement learning proposed by~\citet{zhang2017}.
Due to the nature of the corpus we used for training, only those entries could contribute to the performance of syntactical embellishment that contained two sentences on the simple English, and one on the normal English side.
At the same time, the language employed in the Wikipedia can be expected to be markedly different from the one usually aimed for by storytelling systems, which has an impact on lexical embellishment performance.
Hence, further improvements can be expected from using more---or better suited---corpora, like e.g. Newsela~\cite{xu2015}.
Another promising option would be to explore the viability of embellishment/simplification datasets based on an alignment of simple English versions of novels with their original counterparts.
To exceed sentence-pair based processing, paragraph aligned versions of text would prove to be valuable resources.

Apart from the above theoretical and technological contributions we also presented a first version of Compiled Computer Tales, a corpus of computer generated stories.
This corpus can be used to test the capabilities of embellishment algorithms, as well as for historiographic inquiries.
We cordially invite collaboration to complete its coverage.


\bibliography{lit}

\begin{thebibliography}{26}
\expandafter\ifx\csname natexlab\endcsname\relax\def\natexlab#1{#1}\fi

\bibitem[{Bahdanau et~al.(2014)Bahdanau, Cho, and Bengio}]{bahdanau2014neural}
Dzmitry Bahdanau, Kyunghyun Cho, and Yoshua Bengio. 2014.
\newblock Neural machine translation by jointly learning to align and
  translate.
\newblock \emph{arXiv preprint arXiv:1409.0473}.

\bibitem[{Callaway and Lester(2002)}]{callaway2002}
Charles~B. Callaway and James~C. Lester. 2002.
\newblock Narrative prose generation.
\newblock \emph{Artif. Intell.}, 139(2):213--252.

\bibitem[{Cho et~al.(2014)Cho, Van~Merri{\"e}nboer, Gulcehre, Bahdanau,
  Bougares, Schwenk, and Bengio}]{cho2014learning}
Kyunghyun Cho, Bart Van~Merri{\"e}nboer, Caglar Gulcehre, Dzmitry Bahdanau,
  Fethi Bougares, Holger Schwenk, and Yoshua Bengio. 2014.
\newblock Learning phrase representations using rnn encoder-decoder for
  statistical machine translation.
\newblock \emph{arXiv preprint arXiv:1406.1078}.

\bibitem[{Gerv\'as(2009)}]{gervas2009}
Pablo Gerv\'as. 2009.
\newblock Computational approaches to storytelling and creativity.
\newblock \emph{AI Magazine}, 30(3):49--62.

\bibitem[{Hochreiter and Schmidhuber(1997)}]{hochreiter1997long}
Sepp Hochreiter and J{\"u}rgen Schmidhuber. 1997.
\newblock Long short-term memory.
\newblock \emph{Neural computation}, 9(8):1735--1780.

\bibitem[{Kauchak(2013)}]{kauchak2013}
David Kauchak. 2013.
\newblock Improving text simplification language modeling using unsimplified
  text data.
\newblock In \emph{Proceedings of the 51st ACL}, volume~1, pages 1537--1546.

\bibitem[{Klein et~al.(2017)Klein, Kim, Deng, Senellart, and Rush}]{klein2017}
Guillaume Klein, Yoon Kim, Yuntian Deng, Jean Senellart, and Alexander~M. Rush.
  2017.
\newblock Open{NMT}: Open-source toolkit for neural machine translation.
\newblock \emph{arXiv preprint arXiv:1701.02810}.

\bibitem[{Klein et~al.(1973)Klein, Aeschlimann, and Balsiger}]{klein1973}
Sheldon Klein, J.~Aeschlimann, and D.~Balsiger. 1973.
\newblock Automatic novel writing: A status report.
\newblock \emph{Technical Report 186, Computer Science Department, The
  University of Wisconsin, Madison, Wisconsin}.

\bibitem[{Koehn et~al.(2007)Koehn, Hoang, Birch, {Callison-Burch}, Federico,
  Bertoldi, Cowan, Shen, Moran, and Zens}]{koehn2007}
Philipp Koehn, Hieu Hoang, Alexandra Birch, Chris {Callison-Burch}, Marcello
  Federico, Nicola Bertoldi, Brooke Cowan, Wade Shen, Christine Moran, and
  Richard Zens. 2007.
\newblock {MOSES}: Open source toolkit for statistical machine translation.
\newblock In \emph{Proceedings of the 45th ACL}, pages 177--180. {Association
  for Computational Linguistics}.

\bibitem[{Lebowitz(1985)}]{lebowitz1985story}
Michael Lebowitz. 1985.
\newblock Story-telling as planning and learning.
\newblock \emph{Poetics}, 14(6):483--502.

\bibitem[{Ledig et~al.(2017)Ledig, Theis, Husz{\'a}r, Caballero, Cunningham,
  Acosta, Aitken, Tejani, Totz, Wang et~al.}]{ledig2017photo}
Christian Ledig, Lucas Theis, Ferenc Husz{\'a}r, Jose Caballero, Andrew
  Cunningham, Alejandro Acosta, Andrew~P Aitken, Alykhan Tejani, Johannes Totz,
  Zehan Wang, et~al. 2017.
\newblock Photo-realistic single image super-resolution using a generative
  adversarial network.
\newblock In \emph{CVPR}, volume~2, page~4.

\bibitem[{Luong et~al.(2015)Luong, Pham, and Manning}]{luong2015}
Minh-Thang Luong, Hieu Pham, and Christopher~D. Manning. 2015.
\newblock Effective approaches to attention-based neural machine translation.
\newblock \emph{arXiv preprint arXiv:1508.04025}.

\bibitem[{Meehan(1977)}]{meehan1977tale}
James~R Meehan. 1977.
\newblock Tale-spin, an interactive program that writes stories.
\newblock In \emph{IJCAI}, volume~77, pages 91--98.

\bibitem[{Paetzold and Specia(2017)}]{paetzold2017}
Gustavo~H. Paetzold and Lucia Specia. 2017.
\newblock A survey on lexical simplification.
\newblock \emph{JAIR}, 60:549--593.

\bibitem[{Papineni et~al.(2002)Papineni, Roukos, Ward, and Zhu}]{papineni2002}
Kishore Papineni, Salim Roukos, Todd Ward, and Wei-Jing Zhu. 2002.
\newblock {BLEU}: a method for automatic evaluation of machine translation.
\newblock In \emph{Proceedings of the 40th ACL}, pages 311--318. {Association
  for Computational Linguistics}.

\bibitem[{Pascanu et~al.(2013)Pascanu, Mikolov, and Bengio}]{pascanu2013}
Razvan Pascanu, Tomas Mikolov, and Yoshua Bengio. 2013.
\newblock On the difficulty of training recurrent neural networks.
\newblock In \emph{Proceedings of the 30th ICML}, pages 1310--1318.

\bibitem[{Pennington et~al.(2014)Pennington, Socher, and
  Manning}]{pennington2014}
Jeffrey Pennington, Richard Socher, and Christopher Manning. 2014.
\newblock {GloVe}: Global vectors for word representation.
\newblock In \emph{Proceedings of the 19th EMNLP}, pages 1532--1543.

\bibitem[{Riegl and Veale(2018)}]{riegl2018}
Stefan Riegl and Tony Veale. 2018.
\newblock Live, die, evaluate, repeat: Do-over simulation in the generation of
  coherent episodic stories.
\newblock In \emph{Proceedings of the 9th ICCC}, pages 80--87, Salamanca,
  Spain. {Association for Computational Creativity}.

\bibitem[{Shardlow(2014)}]{shardlow2014}
Matthew Shardlow. 2014.
\newblock A survey of automated text simplification.
\newblock \emph{IJACSA}, 4(1):58--70.

\bibitem[{Siddharthan(2006)}]{siddharthan2006syntactic}
Advaith Siddharthan. 2006.
\newblock Syntactic simplification and text cohesion.
\newblock \emph{Res. Lang. Comput.}, 4(1):77--109.

\bibitem[{Sutskever et~al.(2014)Sutskever, Vinyals, and Le}]{sutskever2014}
Ilya Sutskever, Oriol Vinyals, and Quoc~V. Le. 2014.
\newblock Sequence to sequence learning with neural networks.
\newblock In \emph{Adv Neural Inform Process Syst}, pages 3104--3112.

\bibitem[{Wang et~al.(2016)Wang, Chen, Amaral, and Qiang}]{wang2016}
Tong Wang, Ping Chen, Kevin Amaral, and Jipeng Qiang. 2016.
\newblock An experimental study of {LSTM} encoder-decoder model for text
  simplification.
\newblock \emph{arXiv preprint arXiv:1609.03663}.

\bibitem[{Wen et~al.(2015)Wen, Gasic, Mrksic, Su, Vandyke, and
  Young}]{wen2015semantically}
Tsung-Hsien Wen, Milica Gasic, Nikola Mrksic, Pei-Hao Su, David Vandyke, and
  Steve Young. 2015.
\newblock Semantically conditioned {LSTM}-based natural language generation for
  spoken dialogue systems.
\newblock \emph{arXiv preprint arXiv:1508.01745}.

\bibitem[{Xu et~al.(2015)Xu, {Callison-Burch}, and Napoles}]{xu2015}
Wei Xu, Chris {Callison-Burch}, and Courtney Napoles. 2015.
\newblock Problems in current text simplification research: New data can help.
\newblock \emph{TACL}, 3(1):283--297.

\bibitem[{Zaremba et~al.(2014)Zaremba, Sutskever, and Vinyals}]{zaremba2014}
Wojciech Zaremba, Ilya Sutskever, and Oriol Vinyals. 2014.
\newblock Recurrent neural network regularization.
\newblock \emph{arXiv preprint arXiv:1409.2329}.

\bibitem[{Zhang and Lapata(2017)}]{zhang2017}
Xingxing Zhang and Mirella Lapata. 2017.
\newblock Sentence simplification with deep reinforcement learning.
\newblock In \emph{Proceedings of the 22nd EMNLP}, pages 584--594, Copenhagen,
  Denmark.

\end{thebibliography}
\bibliographystyle{acl_natbib}

\section*{Appendix}

\textbf{Original~\cite{riegl2018} and embellished stories, using the pair-based method:}

\vspace{2mm}
PERSON@1 needed a place to live and PERSON@2 had plenty of it . PERSON@1 found PERSON@2 at an underground lair .
PERSON@1 rented accommodation from her . She paid PERSON@2 what she owed .
LOCATION@1 could not achieve bossy PERSON@1 's lofty goals . She refused to honour PERSON@2 's commitments to her , so PERSON@1 ripped off rich PERSON@2 's best ideas .
PERSON@1 PERSON@2 evicted PERSON@3 from LOCATION@1 's home . At a smoke-filled back room PERSON@3 met PERSON@4 .
LOCATION@1 assiduously curried favor with dictatorial Oscar after cheated PERSON@1 evicted PERSON@2 from LOCATION@2 's home . PERSON@2 told eager Wilde a pack of lies .
PERSON@1 said : '' Dolores wrote propaganda to promote your cause . '' His attitude hardened toward LOCATION@1 .
He openly disrespected PERSON@1 because earlier she took everything that PERSON@2 had . PERSON@1 tried to tune out loudmouthed PERSON@3 's voice .
LOCATION@1 PERSON@1 wrote PERSON@2 off as a loser , so he coldly dismissed PERSON@2 and turned away . It was at the red carpet when PERSON@2 found LOCATION@2 .
PERSON@1 started a new job for influential Rina after unsatisfied PERSON@2 told PERSON@1 to get out and not come back . PERSON@3 took full advantage of her .
She pulled the wool over PERSON@1 's eyes . She said : '' PERSON@2 was a real suck-up to aristocratic PERSON@3 . ''
LOCATION@1 could not reach the bar set by bossy LOCATION@2 . She was very disappointed in her , so '' Get out !
You 're fired '' said PERSON@1 . It was at a recording studio when PERSON@2 found PERSON@3 .
LOCATION@1 PERSON@1 recruited PERSON@2 into her ranks after PERSON@3 asked her to clear out her desk and leave . PERSON@2 took the spotlight from lackadaisical Dolores .
PERSON@1 withheld due payment from lazy Maura . PERSON@2 criticized sinful Dolores in public .
She said : '' PERSON@1 showed no shame in sucking up to influential PERSON@2 . '' She broke with her and went her own way .
What do you think ? Can PERSON@1 and PERSON@2 ever mend their relationship ?
\vspace{2mm}
\hrule
\vspace{2mm}
PERSON@1 needed a place to live and PERSON@2 had plenty of it , and PERSON@1 found PERSON@2 at an underground lair .
PERSON@1 rented accommodation from her and paid PERSON@2 what she owed .
LOCATION@1 could not achieve overbearing PERSON@1 's lofty goals and refused to honor PERSON@2 's commitments to her , so PERSON@1 ripped off rich PERSON@2 's best ideas .
PERSON@1 PERSON@2 evicted from LOCATION@1 's home , and at a smoke-filled back room PERSON@3 met PERSON@4 .
LOCATION@1 assiduously curried favor with dictatorial Oscar after cheated PERSON@1 evicted PERSON@2 from LOCATION@2 's home , who told eager a pack of lies .
PERSON@1 said : '' Dolores wrote propaganda to promote your cause '' .
He openly disrespected PERSON@1 because earlier she took everything that PERSON@2 had , and PERSON@1 tried to tune out loudmouthed PERSON@3 's voice .
LOCATION@1 PERSON@1 wrote PERSON@2 off as a loser , so he coldly dismissed PERSON@2 and turned away at the red carpet when PERSON@2 found LOCATION@2 .
PERSON@1 started a new job for influential Rina after unsatisfied PERSON@2 told PERSON@1 to get out and not come back .
She pulled the wool over PERSON@1 's eyes , saying : '' PERSON@2 was a real suck-up to aristocratic PERSON@3 . ''
LOCATION@1 could not reach the bar set by overbearing LOCATION@2 , and she was very disappointed in her , so '' Get out !
You 're fired '' said PERSON@1 , at a recording studio when PERSON@2 discovered PERSON@3 .
LOCATION@1 PERSON@1 recruited PERSON@2 into her ranks after PERSON@3 asked her to clear out her desk and leave .
PERSON@1 withheld due payment from lazy Maura .
She said : '' PERSON@1 exhibited no shame in digestion up to influential PERSON@2 '' she broke with her and went her own way .
What do you think ? 

\end{document}